# Several categories of Large Language Models (LLMs): A Short Survey

**Saurabh Pahune [1,†,‡], Manoj Chandrasekharan [2]**

[1] Cardinal Health, Dublin OH 43017, USA; Email: saurabh.pahune@cardinalhealth.com, Tel.:+1-901-691-7551
[2] Email: manoj.c@memphis.edu

**Abstract:** Large Language Models (LLMs) have become effective tools for natural language processing and have been used in many different fields. This essay offers a succinct summary of various LLM subcategories. The survey emphasizes recent developments and efforts made for various LLM kinds, including task-based financial LLMs, multilingual language LLMs, biomedical and clinical LLMs, vision language LLMs, and code language models. The survey gives a general summary of the methods, attributes, datasets, transformer models, and comparison metrics applied in each category of LLMs. Furthermore, it highlights unresolved problems in the field of developing chatbots and virtual assistants, such as boosting natural language processing, enhancing chatbot intelligence, and resolving moral and legal dilemmas. The purpose of this study is to provide readers, developers, academics, and users interested in LLM-based chatbots and virtual intelligent assistant technologies with useful information and future directions. This survey sheds light on the possibilities of LLMs and lays the groundwork for additional study and advancement in the area by looking at the background, benefits, and drawbacks of LLMs generally as well as the implications of various LLM models.Thus this paper offers significant information and future directions. Our goal is to look at LLM's history, the advantages and disadvantages of LLMs in general, the types of various LLM models (eg: finance, clinical, multilingual, code, vision), and what all of this implies for the future

---

## 1. Introduction

The origins of the first AI language models can be found in the early history of AI. One of the oldest instances of an AI language model is the ELIZA language model, which made its debut in 1966 at MIT[1,2].An LLM is a development of the language model idea in AI that significantly increases the amount of data utilized for inference and training. As a result, the AI model's capabilities are greatly increased. An LLM normally includes at least one billion parameters, while there isn't a defined size for the data set that must be used for training.A trained deep-learning model called a big language model can read and produce text in a way that is similar to what a human would. Everything is accomplished behind the scenes using a sizable transformer model. In 2017[3] "Attention is All You Need," to establish a transformer model (The 'T' in all the GPT models). It is based on the attention mechanism, dispensing with recurrence and convolutions entirely. Transformer language models use a deep neural network architecture called a Transformer and they are trained to predict either masked words (i.e. fill-in-the-blank) or upcoming words in text[4]. Uszkoreit et al. describe the Transformer, a cutting-edge neural network design based on a self-attention mechanism that aims to be especially effective at interpreting language[5]. Transformer language models have revolutionized the field of natural language processing (NLP) since their introduction in 2018[6]. Transformer language models have received widespread public attention, yet their generated text is often surprising even to NLP researchers[4,7]. As per recent research, some of the top LLMs announced and released in the last few years (e.g. GPT-3/4[8], LLaMA[9], PaLM[10], MiniGPT-4[11], FinGPT[12], OPT[13], BERT[14], Bloomberggpt[15], BLOOM 176B[16], GPT NEO-X[17], RoBERTa [18], Dolly2.0[19] ;)[10,13,20–22]. For applications ranging from web search and chatbots to medical and financial document analysis, many language models are employed in the business[4,15,23].

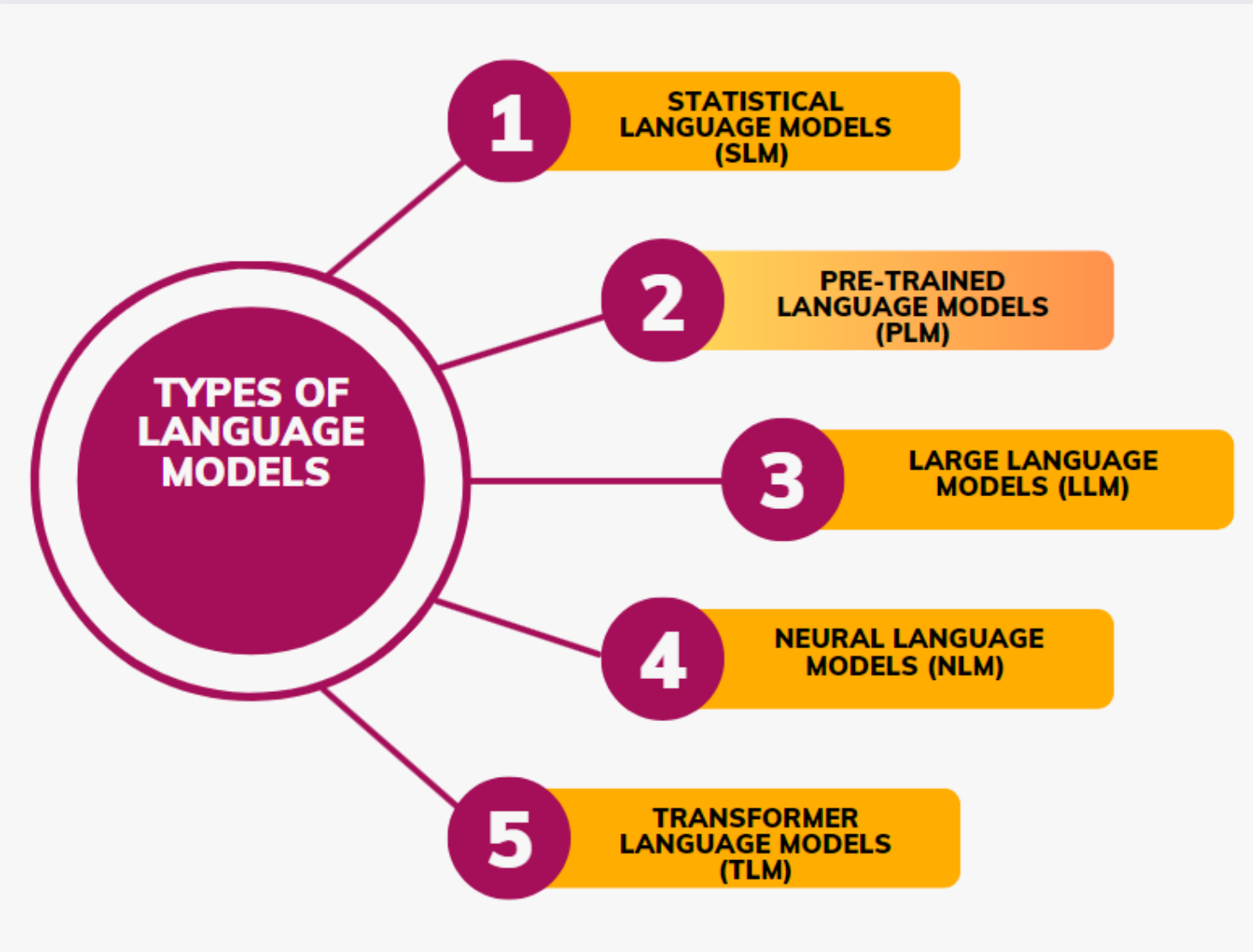

**Figure 1.** Distribution of language models

As per Figure 1. numerous language models have emerged. Language models with a lot of parameters and great processing power are collectively referred to as "Large Language Models" (LLM)[24]. Whereas A sort of language model known as a statistical language model (SLM) uses statistical methods to give probability values to word sequences in a language. It is predicated on the notion that by examining the frequencies and patterns found in a sizable corpus of text, it is possible to predict the likelihood of a specific word appearing in a specific situation[25].In a Neural Language Model (NLM), the probability distribution of word sequences in a language is modeled using neural network topologies. NLMs are made to catch intricate word relationships and produce text that is appropriate for the surrounding context[26,27]. The term "Transformer Language Models" (TLMs) is used to describe language models that especially make use of the Transformer architecture[3]. The term "pre-trained language models" (PLMs) refers to language models that have been trained in an unsupervised fashion on sizable corpora before being adjusted for particular downstream tasks. By extracting patterns and structures from massive volumes of text data, these models learn representations of generic language. In recent years, in the field of healthcare, there are various biomedical and clinical transformer models are available for clinical concept extraction and medical relation[28]. BioBERT[29], ClinicalBERT[30], BioMegatron[31],GatorTron-base[32], GatorTron-medium[32], GatorTron-large [32]. In 2021, One of the largest models in the world for reading comprehension and natural language inference, Megatron-Turing Natural Language Generation 530B was created by Nvidia and Microsoft to facilitate tasks like summarizing and content creation[33]. HuggingFace unveiled BLOOM last year, an open big language model that can produce text in over a dozen programming languages in addition to 46 different natural languages and 13 programming languages[34]. vision language models [35] a family of Visual Language Models (VLM) models that can be rapidly adapted to novel tasks using only a handful of annotated examples is an open challenge for multimodal machine learning research. Table1. cited the work as having been mentioned by the corresponding author on this language model field.

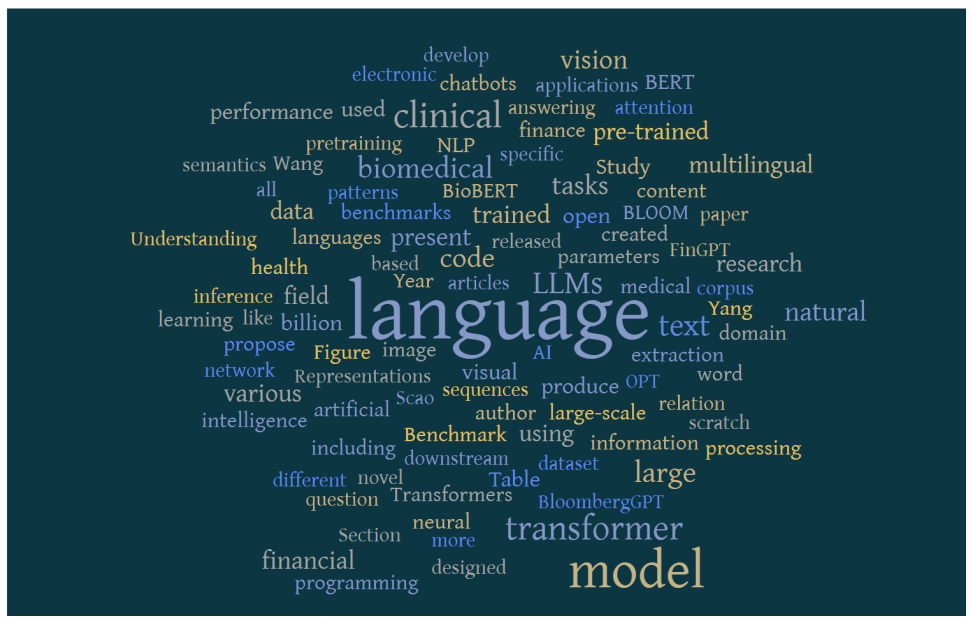

**Figure 2.** A word cloud showing the frequency of terms used in the articles we reviewed.

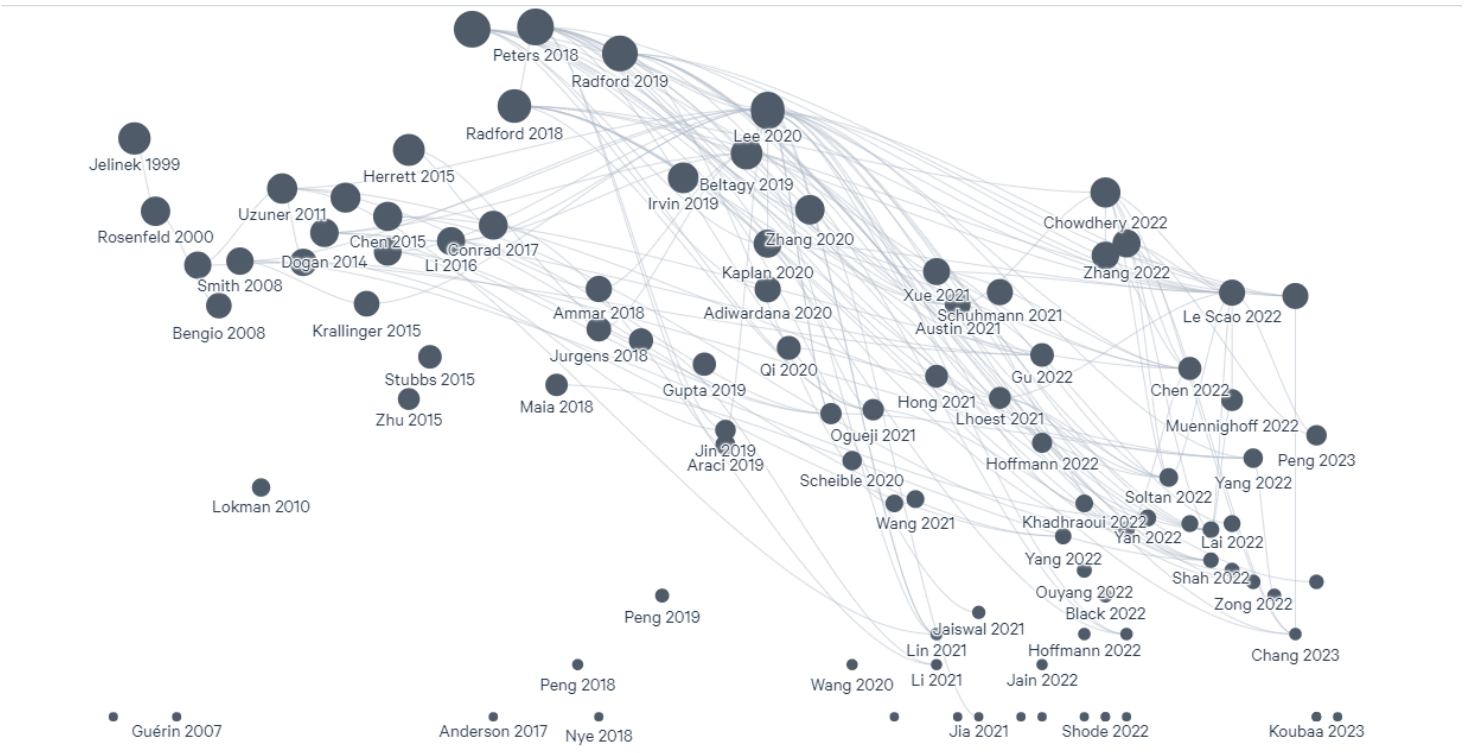

**Figure 3.** Mapping of visualize research articles

In Figure 3., by using connecting lines to visually represent the citations for this work, it demonstrates the connection. Useful research resources and project state are included in this time-based citation network to arrange this review of the literature. The cited research addresses several topics, including model features, datasets, transformer models, and benchmarks for assessing LLM performance.

The rest of the article is organized as follows. Section 2.1 presents prior related work that has various versions of LLM models (Finance base, Clinical base, Vision base, Code-base, Multilingual based)and tables respectively. In Section 3, define the configuration of various models with billions of parameters and created taxonomy tables to discuss the detailed methodology of LLM models such as (Benchmark and dataset, Dataset content, Implementation details etc.), A mapping of the articles used in this paper was also developed Figure 3. In Section 4 (Open Issues and Research Directions ), the open issues and potential future directions of LLM mdoels. The conclusions are described in Section 5.

**Table 1.** Studies on different language models

| Types of Language Models | Study |
|---|---|
| Statistical language models (SLM) | Jelinek et al. [36], Rosenfeld et al. [37] |
| Pre-trained language models (PLM) | Matthew et al.[38] |
| Large language models (LLM) | Kaplan et al.[39] |
| Neural language models (NLM) | Bengio et al.[40] |
| Transformer language models (TLM) | Vaswani et al.[3] |

## 2. RELATED WORK

### *2.1. Multilingual language-image model*

An artificial intelligence model that can comprehend and produce both textual and visual content across several languages is called a multilingual language-image model. In order to assess and provide useful information from both modalities, these models are particularly trained to process and understand the complicated relationships between words and visuals. Chen et al. [41] study PaLI (Pathways Language and Image model), a model that extends this approach to the joint modeling of language and vision. Scheible et al.[42] discovered GottBERT is a pure German Language Model. AlexaTM 20B Soltan at al.[43] demonstrate that multilingual large-scale sequence-to-sequence (seq2seq) models,).pre-trained on a mixture of denoising and Causal Language Modeling (CLM). Dossou et al.[44] present AfroLM, a multilingual language model pretrained from scratch on 23 African languages (the largest effort to date) using our novel self-active learning framework. Scao et al.[16] present BLOOM, a 176B-parameter open-access Multilingual language model designed.

**Table 2.** Various multilingual language LLMs.

| Models | Study | Year |
|---|---|---|
| BLOOM | Scao et al.[16] | 2022 |
| AfroLM | Dossou et al.[44] | 2022 |
| AlexaTM 20B | Soltan at al.[43] | 2022 |
| GottBERT | Scheible et al.[42] | 2020 |
| PaLI | Chen at al.[41] | 2022 |

### *2.2. Clinical and biomedical transformer model*

A clinical and biomedical transformer model is a type of artificial intelligence that was created with the express purpose of processing and analyzing clinical and biomedical text data. These transformer models make use of the architecture of the transformer, which has excelled in jobs requiring natural language processing. Clinical notes, electronic health records, research articles, and other pertinent sources of clinical and biomedical data are included in the large-scale datasets used to train the clinical and biomedical transformer models. These models gain knowledge of the specific words, phrases, and ideas used in the medical field. Clinical and biomedical transformer models' main goals are to derive insightful information, carry out text categorization, entity recognition, relation extraction, question answering, and other activities particular to the clinical and biomedical area. Clinical decision support, information retrieval, patient risk assessment, and automated documentation are just a few of the activities that they may help healthcare workers with. Yang et al.[32] develop from scratch a large clinical language model – GatorTron – using >90 billion words of text (including >82 billion words of de-identified clinical text) and systematically evaluate it on 5 clinical NLP tasks including clinical concept extraction, medical relation extraction, semantic textual similarity, natural language inference (NLI), and medical question answering (MQA). Lee et al.[29] introduce BioBERT (Bidirectional Encoder Representations from Transformers for Biomedical Text Mining), which is a domain-specific language representation model pre-trained on large-scale biomedical corpora. Li et al[45] Present Hi-BEHRT, a hierarchical Transformer-based model that can significantly expand the receptive field of Transformers and extract associations from much longer sequences. Using a multimodal large-scale linked longitudinal electronic health records. Wang et al.[46] propose an innovative causal inference model–InferBERT, by integrating the A Lite Bidirectional Encoder Representations from Transformers (ALBERT). Large language models in health care As per Anmol et al.[47] has already been proposed that LLMs, such as ChatGPT, could have applications in the field of health care due to the large volumes of free-text information available for training models.An LLM trained on more than 90 billion words of text from electronic health records (EHR)[28] Author Yang et al. develop a scratch a large clinical language model GatorTron using more than 90 billion words of text. Existing biomedical and clinical transformer models for clinical concept extraction and medical relation such as BioBERT[29], ClinicalBERT[30], BioMegatron[31],GatorTron-base[32], GatorTron-medium[32], GatorTron-large [32].

Santosh et al.[48] propose PathologyBERT - a pre-trained masked language model which was trained on 347,173 histopathology specimen reports and publicly released in the Huggingface1repository. Comprehensive experiments demonstrate that pre-training of transformer model on pathology corpora yields performance improvements on Natural Language Understanding (NLU) and Breast Cancer Diagnose Classification when compared to nonspecific language models. Jaiswal et al.[49] intorduce RadBERT-CL which is"Factually-Aware Contrastive Learning For Radiology Report Classification." Also show that the representations learned by RadBERT-CL can capture critical medical information in the latent space. Gu et al.[14] accelerate research in biomedical and released state-of-the-art pretrained and task-specific models for the community, and created a leaderboard featuring BLURB benchmark (Biomedical Language Understanding Reasoning Benchmark)). The author challenges, the major advantage of domain-specific pretraining from scratch stems from having an in-domain vocabulary. Peng et al.[50] introduce BLUE, a collection of resources for evaluating and analyzing biomedical natural language representation models. find that the BERT models pre-trained on PubMed abstracts and clinical notes see better performance than do most state-of-the-art models. Beltagy et al.[51] SCIBERT leverages unsupervised pretraining on a large multi-domain corpus of scientific publications to improve performance on downstream scientific NLP tasks. Alsentzer et al.[30] released Clinical BERT models for clinical text: one for generic clinical text and another for discharge summaries specifically. Also, demonstrate on several clinical NLP tasks that improvements this system offers over traditional BERT and BioBERT. Shin et al.[31] come up with BioMegatron consider as large biomedical domain lanuage model. Which show consistent improvements on benchmarks with larger BioMegatron model trained on a larger domain corpus, contributing to our understanding of domain language model applications.

**Table 3.** Various biomedical and clinical LLMs.

| Models | Study | Year |
|---|---|---|
| GatorTron-base | Yang et al.[32] | 2022 |
| BioBERT | Lee et al.[29] | 2020 |
| EntityBERT | Lin et al.[52] | 2021 |
| Hi-BEHRT | Li et al[45] | 2022 |
| InferBERT | Wang et al.[46] | 2021 |
| PathologyBERT | Santosh et al.[48] | 2022 |
| PubMedBERT | Gu et al.[14] | 2021 |
| SciBERT | Beltagy et al.[51] | 2019 |
| RadBERT | Yan et al.[53] | 2022 |
| ClinicalBERT | Alsentzer et al.[30] | 2019 |
| BlueBERT | Peng et al.[50] | 2019 |
| BioMegatron | Shin et al.[31] | 2019 |

### *2.3. Large language model for finance*

These models can analyze and grasp complicated financial text data efficiently by making use of deep learning techniques like transformer architectures. They can help with jobs including compiling financial reports, summarizing financial documents, researching investments, managing portfolios, and analyzing financial news. Financial professionals' ability to make more educated, data-driven decisions may be improved by the use of large language models in the field. They can offer insights for investing plans, assist in identifying market trends, evaluate risk factors, and spot abnormalities. Wu et al.[15] present BloombergGPT (A large language model for finance), a 50 billion parameter language model that is trained on a wide range of financial data. Author validates BloombergGPT on standard LLM benchmarks, open financial benchmarks, and a suite of internal benchmarks that most accurately reflect our intended usage. Scao et al.[16] present BLOOM, a 176B-parameter open-access language model designed and built. BLOOM is a decoder-only Transformer language model that was trained on the ROOTS corpus, a dataset comprising hundreds of sources in 46 natural and 13 programming languages (59 in total). Black et al.[17] introduce GPT-NeoX-20B, a 20 billion parameter autoregressive language model trained on the Pile[54],evaluate its performance on a range of language-understanding, mathematics, and knowledge-based tasks. Araci et al.[55]introduce FinBERT language model based on BERT, to tackle NLP tasks in the financial domain. Zhang et al.[13] present Open Pre-trained Transformers (OPT), a suite of decoder-only pre-trained transformers ranging from 125M to 175B parameters. Yang et al.[12] present an open-source large language model FinGPT, for the finance sector. FinGPT responds innovatively by leveraging pre-existing LLMs and fine-tuning them to specific financial applications. Xie at al.[56] discovers the PIXIU LLM model for Instruction Data and Evaluation Benchmark for Finance.

**Table 4.** Various finance-based LLMs.

| Models | Study | Year |
|---|---|---|
| BloombergGPT | Wu et al.[15] | 2023 |
| GPT-NeoX | Black et al.[17] | 2022 |
| OPT | Zhang et al.[13] | 2022 |
| BLOOM-176B | Scao et al.[16] | 2022 |
| FinBERT | Araci et al.[55] | 2019 |
| FinGPT | Yang et al.[12] | 2023 |
| PIXIU | Xie at al.[56] | 2023 |

### *2.4. Classifications of vision language models*

Artificial intelligence models called "vision language models" are created to comprehend and produce data from the combination of visual and linguistic inputs. These models seek to close the gap between the comprehension of images or other visual content and that of natural language. These models are capable of carrying out a number of tasks, such as image captioning, visual question answering (VQA), image generation from text descriptions, and image-text matching. For instance, a vision language model can produce a caption for an image that accurately describes the image's content. Similar to this, the model can offer pertinent responses or justifications when asked a text query regarding a picture. Alayrac et al.[57] comes with Flamingo, a family of Visual Language Models (VLM) models that can be rapidly adapted to novel tasks using only a handful of annotated examples is an open challenge for multimodal machine learning research. Monajatipoor et al.[58] Vision-and-language (VL) models take image and text as input and learn to capture the associations between them.Prior studies show that pre-trained VL models can significantly improve the model performance for downstream tasks such as Visual Question Answering (VQA). Ko et al.[59] present F-VLM, a simple open-vocabulary object detection method built upon Frozen Vision and Language Models.Where F-VLM simplifies the current multi-stage training pipeline by eliminating the need for knowledge distillation or detection tailored pretraining. Zhu et al.[11] projected MiniGPT-4 which is a enhancing vision-language understanding with advanced large language models. Hong et al.[60] propose a recurrent BERT model that is time-aware for use in of vision-and-language navigation(VLN). In this paper author propose a recurrent BERT model that is time-aware for use in VLN. Thrush et al.[61] present a novel task and dataset for evaluating the ability of vision and language models to conduct visio-linguistic compositional reasoning, which we call Winoground. Wang et al.[62] propose a smaller and faster VL model, MiniVLM, which can be finetuned with good performance on various downstream tasks like its larger counterpart. MiniVLM consists of two modules, a vision feature extractor and a transformer-based vision-language fusion module.

**Table 5.** Various vision language LLMs.

| Models | Study | Year |
|---|---|---|
| Flamingo | Alayrac et al.[57] | 2022 |
| BERTHop | Monajatipoor et al.[58] | 2022 |
| F-VLM | Kuo et al.[59] | 2022 |
| MiniVLM | Wang et al.[62] | 2020 |
| VLN-BERT | Hong et al.[60] | 2021 |
| Winoground | Thrush et al.[61] | 2022 |
| MiniGPT-4 | Zhu et al.[11] | 2023 |

### 2.5. *Classifications of code large language model (Code LLMs)*

Large-scale language models have the potential to promote programming by promoting code reuse, knowledge sharing, and developer collaboration. They can aid in eliminating errors, automating repetitive coding processes, and accelerating the development process. A code big language model is designed to help programmers with a variety of coding-related tasks. These models are capable of tasks like code completion, code generation, code summarization, and code translation and can comprehend the syntax, semantics, and programming patterns of code. Luo et al.[63] In this paper introduced WizardCoder, which empowers Code LLMs with complex instruction fine-tuning, by adapting the Evol-Instruct method to the domain of code. Nijkamp et al.[64] release a family of large language models up to 16.1B parameters, called CODEGEN, on natural language and programming language data. Jain et al.[65] present an approach to augment these large language models with post-processing steps based on program analysis and synthesis techniques, that understand the syntax and semantics of programs. Wang et al. [66] present CodeT5,a unified pre-trained encoder-decoder Transformer model that better leverages the code semantics conveyed from the developer-assigned identifiers.

**Table 6.** Various code language models (Code LLMs)

| Models | Study | Year |
|---|---|---|
| WizardCoder | Luo et al.[63] | 2023 |
| CodeGen | Nijkamp et al.[64] | 2022 |
| Jigsaw | Jain et al.[65] | 2022 |
| CodeT5 | Wang et al. [66] | 2021 |

## 3. Taxonomy tables of various LLM models

**Table 7.** Classifications of multilingual language model

| LLM Model | Benchmark and Dataset | Dataset content | Implementation details | Application | Versions of Model |
|---|---|---|---|---|---|
| PALI [41] | WebLI[67],COCO[68], Multilingual captioning on Crossmodal-3600[69], VQA[70] | New image-text training set containing 10B images and texts in over 100 languages | - | Pathways Language and Image model | PaLI-3B, PaLI-15B, PaLI-17B |
| GottBERT[42] | The German data portion of the OSCAR measures 145GB of text containing approximately 21.5 billion words in approximately 459 million documents (one document per line) | A vocabulary of 52k subword tokens based on 40 GB of randomly sampled documents of the German OSCAR portion. | 256 core TPU pod using the RoBERTa BASE architecture | A pure German Language Model | GottBERT, German BERT, XLM RoBERTa, dbmz BERT[42] |
| BLOOM [16] | BLOOM is a decoder-only Transformer language model that was trained on the ROOTS corpus[71], 2022), A composite collection of 498 Hugging Face datasets[72], HELM benchmark[73] | Dataset comprising hundreds of sources in 46 natural and 13 programming languages (59) | 156 TFLOPs in our fastest configuration with NVIDIA V100 GPUs | A 176B-Parameter Open-Access Multilingual Language Model | BLOOM-1B7, BLOOM-560M, BLOOM-1.1B, BLOOM-1.7B, BLOOM-1.7B, BLOOM-7.1B |

Table 7 – continued from previous page

| | | | | | | |
|---|---|---|---|---|---|---|
| AlexaTM 20B[43] | Wikipedia and mC4 [70] | Data in 12 languages, namely, Arabic, English, French, German, Hindi, Italian, Japanese, Marathi, Portuguese, Spanish, Tamil, and Telugu, Pack sequences of tokens to produce sequences of approximately 1024 subword units. | Trained AlexaTM 20B for 120 days on 128 A100 GPUs | A Large-Scale Multilingual Seq2seq Model | - | - |
| AfroLM[44] | Languages Corpora Details [64,74,75], YOSM dataset[76] | Several works on sentiment analysis have been done on high resource languages while low resources languages like Yoruba and other African languages, data comprised 1500 movie reviews that were sourced from IMDB, Rotten Tomatoes, Letterboxd, Cinemapointer, and Nollyrated[76] | Google Cloud with a single 48GB NVIDIA A100 GPU | A Self-Active Learning-based Multilingual Pre-trained Language Model for 23 African Languages | AfroLM-Large, AfroLM-Large (w/AL) to AfroLM-Large (w/o AL), AfriBERTa-Large[44] | - |

**Table 8.** Classifications of various clinical/biomedical transformer models

| LLM Model | Benchmark and Dataset | Dataset content | Implementation details | Application | Versions of Model |
|---|---|---|---|---|---|
| GatorTron-base [32] | 5 clinical NLP tasks named entity recognition [NER], medical relation extraction (MRE), semantic textual similarity (STS), natural language inference(NLI), and medical question answering (MQA), MedNLI[77], 2019 n2c2[78], emrQA Medication [79] | A total number of 290,482,002 clinical notes from 2,476,628 patients were extracted from the UF Health Integrated Data Repository (IDR), the enterprise data warehouse of the UF Health system[32] | 992 A100 80G GPUs from 124 NVIDIA DGX | A large clinical language model | GatorTron-base, GatorTron-base(1/4 data),GatorTron-medium, GatorTron-large |
| BioBERT [29] | Pre-trained on biomedical domain corpora (PubMed abstracts, English Wikipedia, BooksCorpus and PMC full-text articles) | NCBI Disease with 6881 (Number of annotations),2010 i2b2/VA disease with 19665 (Number of annotations) , BC5CDR disease with 12694 (Number of annotations) | Eight NVIDIA V100 GPUs. | A pre-trained biomedical language representation model for biomedical text mining | BioBERT v1.0, BioBERT v1.1 |
| Hi-BEHRT [45] | Model performance on incident risk prediction for four diseases: heart failure (HF), diabetes, chronic kidney disease (CKD), and stroke. | Clinical Practice Research Datalink (CPRD)[80], myocardial failure[81], diabetes mellitus [82] | 2 GPUs [45] | Hierarchical Transformer-based model for accurate prediction of clinical events using multimodal longitudinal electronic health records | BEHRT, Hi-BEHRT |
| InferBERT [46] | Two FAERS datasets, Analgesics-induced acute liver failure and Tramadol-related mortalities datasets | FAERS dataset Effects: (Acute liver fibrosis and cirrhosis, OR Acute liver failure and associated disorders, OR Cholestasis and jaundice) AND Drugs by AND-groups: [Analgesics (Any Role)]" to extract 45,773 [83] | one NVIDIA Tesla V100 GPU. | A Transformer-Based Causal Inference Framework for Enhancing Pharmacovigilance | - |

Table 8 – continued from previous page

| | | | | | |
|---|---|---|---|---|---|
| PathologyBERT[48] | Two non-overlapping corpora of histopathology reports from Emory University Hospital (EUH) | Emory University Institutional Review Board(IRB), a total of 340,492 unstructured Histopathology specimens from 67,136 patients were extracted from the clinical data warehouse of Emory University Hospital (EUH) between the years 1981 and 2021. | GeForce Quadro RTX 6000 24 GB GPUs | A New Transformer Language Model for Pathology Domain(Note: PathologyBERT on Corpus II to predict 6 breast cancer diagnose severity (invasive breast cancer, high risk lesion, borderline lesion, non-breast cancer, benign, and negative)[48]. | - |
| RadBERT [53] | MIMIC-CXR dataset [84], pseudo-labeled using automatic labeler [85] | MIMIC-CXR dataset which consists of 377, 110 chest-Xray images of 227, 827 patients along with their corresponding de-identified radiology reports. | - | Factually-Aware Contrastive Learning For Radiology Report Classification | RadBERT-CL |
| PubMedBERT [14] | BLURB: A Comprehensive Benchmark for Biomedical NLP, BC5-chem[86],BC5-disease[86], NCBI-disease[87], BC2GM[88], PubMedQA[89] | BLUE mixes PubMed-based biomedical applications (six datasets such as BC5, ChemProt, and HoC) with MIMIC-based clinical applications (four datasets such as i2b2 and MedNLI) | One DGX-2 machine with 16 V100 GPUs | It was pretrained over PubMed abstracts and complete PubMed Central articles and is an uncased BERT Base model | BERT, RoBERTa, BioBERT, SciBERT, Clinical- BERT , and BlueBERT |
| SciBERT[51] | Corpus, EBM-NLP[90], SciERC[91], ACL-ARC[92] | Train SCIBERT on a random sample of 1.14M papers from Semantic scholar[93] corpus consists of 18% papers from the computer science domain and 82% from the broad biomedical domain | Single TPU v3 with 8 cores | A Pretrained Language Model for Scientific Text | - |
| ClinicalBERT[30] | i2b2 2006[94], i2b2 2010[95], i2b2 2012[96], i2b2 2014[97] | Clinical NLP Task, MedNLI natural language inference task[98], Use clinical text from the approximately 2 million notes in the MIMIC-III v1.4 database[99] | A single GeForce GTX TITAN X 12 GB GPU | Evaluates representations of clinical notes using bidirectional transformers (ClinicalBERT) | Bio+Clinical BERT, Clinical BERT [30] |

Table 8 – continued from previous page

| | | | | | |
|---|---|---|---|---|---|
| BioMegatron[31] | Used downstream biomedical benchmark datasets for Named Entity Recognition(NER), Relation Extraction(RE), and Question Answering(QA) | BC5CDR[86] NER dataset annotated disease, ChemProt[100] dataset contains sentences from PubMed abstracts, BioASQ-7b factoid task[101] is a biomedical QA dataset | Batch size of 64 per GPU with data parallelism on 16 GPUs | BioMegatron consider as large biomedical domain language model | BioMegatron-345m, BioMegatron-800m, BioMegatron-1.2b |
| BlueBERT[50] | Uses both PubMed text and de-identified clinical notes from MIMIC-III[99], PubMed abstract[102] | Models were trained with 5M steps on the PubMed corpus and 0.2M steps and BLUE contains five tasks with ten corpora that cover a broad range of data quantities and difficulties. (Sentence similarity, Named entity recognition, Relation extraction, Document classification, Inference task ) | - | Biomedical Language Understanding Evaluation (BLUE) benchmark to facilitate research in the development of pre-training language representations in the biomedicine domain | BlueBERT-Base + Uncased +PubMed, BlueBERT-Base + Uncased+PubMed + MIMIC-III |

**Table 9.** Classifications of large language model for finance

| LLM Model | Benchmark and Dataset | Dataset content | Implementation details | Application | Versions of Model |
|---|---|---|---|---|---|
| BloombergGPT[15] | C4[103], FinPile[54], public financial NLP benchmarks [104] | Colossal Clean Crawled Corpus (C4) gives us a vocabulary size of 125,000, Dump of English Wikipedia from July 1, 2022. | 8 NVIDIA 40GB A100 GPUs | A large language model for finance | BLOOM-style,BLOOM176B[16] |
| GPT-NeoX [17] | Pile [54] | It has 22 data sources, coarsely broken down into 5 categories (Academic Writing, Web-scrapes and Internet Resources, Prose, Dialogue, Miscellaneous) | 8 NVIDIA A100-SXM4-40GB GPUs and configured with two AMD EPYC 7532 CPUs | An Open-Source Autoregressive Language Model | GPT-NeoX-20B |
| OPT-175B[13] | BookCorpus[105], MinhashLSH[106], RoBERTa CCNews[18] | Eight Transformer language models ranging from 125 million to 175 billion parameters | On 992 80GB A100 GPUs | Open Pre-trained Transformer Language Models | OPT from 125M to 175B (Ex: OPT-125M to OPT-175B) |
| BLOOM-176B[16] | ROOTS corpus[71], A composite collection of 498 Hugging Face datasets[72], SuperGLUE[107], STS datasets from MTEB[108], HELM benchmark[73] | ROOTS corpus (dataset comprising hundreds of sources in 46 natural and 13 programming languages (59 in total)) | 8 NVIDIA A100 80GB GPUs | A 176B-Parameter Open-Access Multilingual Language Model | BLOOM-560M,BLOOM-1B7, BLOOM-1.7B, BLOOM-3B,BLOOM-7.1B, BLOOMZ[109] |
| FinBERT[55] | Financial corpus(TRC2-financial)[110], Financial PhraseBank[111], FiQA Sentiment[112] | FiQA Sentiment is a dataset that was created for financial opinion mining and question answering challenge, use the data for Task 1, which includes 1,174 financial news headlines and tweets with their corresponding sentiment score | Amazon p2.xlarge EC2 instance with one NVIDIA K80 GPU, 4 vCPUs | Financial sentiment analysis with pre-trained language models | FinBERT-task, FinBERT-domain |
| FinGPT[12] | Academic datasets, Novel financial dataset | Different financial data sources, such as Financial News, Company Fillings, Social Media Discussions, and Company Announcements | - | An open-source large language model (Financial sentiment analysis, Financial Fraud detection, Credit scoring, Portfolio optimization, Financial education) | FinLLM |

Table 9 – continued from previous page

| LLM Model | Benchmark and Dataset | Dataset content | Implementation details | Application | Versions of Model |
|---|---|---|---|---|---|
| PIXIU [56] | Financial Evaluation Benchmark, financial instruction tuning dataset FIT on various financial NLP, FiQA-SA, Financial Phrase Bank (FPB), Benchmark FLARE | FiQA-SA consist 11,730 instructions (news headlines,tweets), FPB has 48,450 news data types [56] | 8 A100 40GB GPUs. | A Large Language Model, Instruction Data and Evaluation Benchmark for Finance | FinMA |

**Table 10.** Classifications of vision language models

| LLM Model | Benchmark and Dataset | Dataset content | Implementation details | Application | Versions of Model |
|---|---|---|---|---|---|
| Flamingo[57] | MultiModal MassiveWeb (M3W) dataset, Pairs of image/video and text, ALIGN [113], VQAv2, VATEX, VizWiz | ALIGN dataser composed of 1.8 billion images paired with alt-text, LTIP (Long Text and Image Pairs) which consists of 312 million image and text pairs. | - | A Visual Language Model for Few-Shot Learning | Flamingo-3B, Flamingo-9B, Flamingo |
| BERTHop [58] | ChestX-ray14, MIMIC-CXR | MIMIC-CXR labels are generated using ChexPert[85] and NegBio[114] auto labelers. OpenI comprises 3,996 reports and 8,121 associated images from 3,996 unique patients collected by Indiana University from multiple institutes. | - | An Effective Vision-and-Language Model for Chest X-ray Disease Diagnosis | PixelHop++, BlueBERT |
| F-VLM [59] | LVIS Benchmark[115] | LVIS dataset which contains a large and diverse set of 1203 object categories suitable for open-vocabulary detection. | - | A simple open-vocabulary object detection method built upon Frozen Vision and Language Models. | F-VLM-R50, F-VLM-R50x4,F-VLM-R50x16,F-VLM-R50x64 |
| MiniGPT-4 [11] | Conceptual Caption[116], SBU[117] and LAION[118] | Conceptual 12M (CC12M), a dataset with 12 million image-text pairs specifically meant to be used for vision-and-language pre-training | 4 A100 (80GB) GPUs | MiniGPT-4 aims to align visual information from a pretrained vision encoder with an advanced large language model (LLM) | – |
| MiniVLM [62] | Image Captioning[119], VQA[120], Natural Language Visual Reasoning for Real (NLVR2)[121], COCO image captioning task | VQA is a representation of [CLS] is used to predict the answer over a shared set of 3129 answers with a linear layer | 2 Intel(R) Xeon(R) CPU E5-2620 v4 @2.10GHz | A Smaller and Faster Vision-Langauge Model | - |
| VLN-BERT[60] | R2R[122], REVERIE[123] | Matterport3D, a large-scale RGB-D dataset containing 10,800 panoramic views from 194,400 RGB-D images of 90 building-scale scenes[124] | Single NVIDIA 2080Ti GPU | A Recurrent Vision-and-Language BERT for Navigation | VisualBERT, VL BERT |

Table 10 – continued from previous page

| | | | | | |
|---|---|---|---|---|---|
| Winoground[61] | Winoground dataset was hand-curated by four expert annotators with extensive experience in vision and language research as well as computational linguistics | Dataset has 1600 image-text pairs in total, with 800 correct and 800 incorrect pairings | - | Probing Vision and Language Models for Visio-Linguistic Compositionality | - |

**Table 11.** Classifications of code language models (Code LLMs)

| LLM Model | Benchmark and Dataset | Dataset content | Implementation details | Application | Versions of Model |
|---|---|---|---|---|---|
| WizardCoder [63] | Four code generation benchmarks: HumanEval[125], HumanEval+[126], MBPP[127], and DS-1000[128] | Initialized it with the 20K instruction-following dataset called Code Alpaca[67]. Evol-Instruct technique on this dataset consisting of 20,000 samples to produce evolved data. | - | Empowering Code Large Language Models with Evol-Instruct | WizardLM [129] |
| CodeGen [64] | THEPILE, BIGQUERY, and BIGPYTHON. | THEPILE is an 825.18 GiB English text corpus collected [54], BIGQUERY is a subset of Google's publicly available BigQuery dataset, dataset BIGPYTHON contains a large amount of data in the programming language, Python. | Google's TPU-v4 hardware | An open LLM for code with multi-turn program synthesis | CodeGen -MULTI , CodeGen-NL, CodeGen-MONO |
| Jigsaw [65] | PandasEval dataset, Hackathon dataset | This dataset consists of 68 Python Pandas tasks. Each task can be solved using a single line of code by composing at most 2-3 Pandas functions, Hackathon dataset consists of 21 Pandas tasks; each task | - | Large Language Models meet Program Synthesis | - |
| CodeT5[66] | SearchNet[130] | Consists of 99 natural language queries with about 4k expert relevance annotations of likely results from CodeSearchNet Corpus. The corpus contains about 6 million functions from open-source code spanning six programming languages (Go, Java, JavaScript, PHP, Python, and Ruby) | 16 NVIDIA A100 GPUs with 40G memory. | Unified Pre-trained Encoder-Decoder Models for Code Understanding and Generation | CodeBERT |

**Table 12.** Detailed of several existing LLMs configuration with Millions/ Billions of parameters.

| Model | Optimizer and Layers | Model size | Reference |
|---|---|---|---|
| GPT-2 | Adam, 12 layers | 1.5 billion | [131] |
| GPT-3 | Adam, 96 layers | 175 billion | [132,133] |
| Microsoft DialoGPT | - | 147 million | [134,135] |
| BloombergGPT | GELU, 70 layers | 50 billion | [15] |
| Vicuna | - | 13 billion | [136] |
| Dolly2.0 | - | 12 billion | [19] |
| BLOOM | Adam,70 layers | 176 billion | [34] |
| LLaMA | AdamW,- | 65 billion | [9] |
| Jurassic-1 | - | 178 billion | [137] |
| GLM | AdamW,- | 130 billion | [138] |
| PaLM | Adafactor,- | 540 billion | [139] |
| OPT 175B | AdamW, 96 | 175 billion | [13] |
| Chinchilla | Adam,80 layers | 70 billion | [67] |
| BERT-base | Adam,12 layers | 100 million | [14] |
| BERT-large | Adam,24 layers | 300 million | [14] |
| ALBERT | Adam ,12 layers | 12 million | [140] |
| RoBERTa base | Adam,12 layers | 125 million | [18] |
| RoBERTa large | Adam,24 layers | 355 million | [18] |
| Megatron-Turing NLG | - | 530 billion | [141] |
| BioBERT | - | 13.5 billion | [29] |
| ClinicalBERT | Adam | 1.28 billion | [30,142] |
| BioMegatron | Adam,24 | 1.2 billion | [31] |
| GatorTron-base | Adam,24 layers | 345 million | [32,143] |
| GatorTron-medium | Adam,48 layers | 3.9 billion | [32,143] |
| GatorTron-large | Adam, 56 layers | 8.9 billion | [32,143] |
| Gopher | Adam,- | 280 billion | [144] |
| GPT-NeoX | AdamW | 20 billion | [17] |
| Bloom 176 | Adam,24 layers | 176 billion | [16] |
| PubMedBERT | - | 110 million | [89] |
| AlexaTM 20B | -,46layers | 19.75 billion | [43] |
| AfroLM-Large | -,10layers | 264 million | [44] |
| Hi-BEHRT | Adam, layers | 264 million | [45] |
| PathologyBERT | Adam, 12 Layers | 347 million | [48] |
| BioMegatron | Adam, 24 Layers | 345 million | [31] |
| BioMegatron medium | Adam, 36 Layers | 800 million | [31] |
| BioMegatron large | Adam, 24 Layers | 1.2 billion | [31] |
| BloombergGPT | Adam, 70 Layers | 50.6 billion | [15] |
| BLOOM-style | Adam, 70 Layers | 50 billion | [145] |
| GPT-NeoX-20B | Adam, 44 Layers | 20 billion | [17] |
| CODEGEN | - | 16.1 billion | [64] |

In Table12. based on what we've seen, the billions to millions range. Dataset optimization is a crucial step in LLM models, particularly those with a large number of parameters, with the goal of improving the model's functionality and speed. To make sure the training data is representative, diverse, and in line with the anticipated results, dataset optimization entails carefully choosing and preparing the training data. Researchers and programmers can enhance the model's capacity to comprehend and produce words, leading to more precise and cogent responses, by optimizing the dataset. Basically, dataset optimization helps LLM models reach their full potential by supplying high-quality training data that is in line with the particular tasks or objectives at hand.

## 4. Open Issues and Research Directions

Due to the size of large language models, their deployment requires a high level of technical expertise, including a firm understanding of deep learning, transformer models, distributed software, and hardware as well as ethical and legal issues arising from the liability and harm potential of such systems.

Many professionals in the IT sector are working to support research and create technologies that can open up access to broad language models, allowing customers and companies of all sizes to take advantage of them.

It is not clear how large clinical language models with billions of parameters can help medical AI systems utilize unstructured electronic health records (EHRs) within the current legal and ethical framework while ensuring privacy of patient information and accuracy of the information provided[28].

Scaling and maintaining large language models can be difficult and expensive. Building a foundational large language model often requires months of training time and millions of dollars [33].

And because LLMs require a significant amount of training data, developers and enterprises can find it a challenge to access large-enough datasets to train such systems while ensuring data is collected ethically and with permission of the parties involved. Maintaining them by putting in place systems to ensure accurate and useful outputs at scale is also a significant challenge.

## 5. Conclusion

In this study, the most recent advances in large language models (LLMs) were showcased and the key concepts, findings, and strategies for understanding and exploiting LLMs were presented. A wide range of issues are covered in this study, including model features, datasets, transformer models, and LLM performance benchmarks. Recent studies have focused on various LLM types, such as multilingual LLMs, biomedical and clinical LLMs, vision language LLMs, and code language models. This survey attempts to cover the most recent research on LLMs and provides academics and engineers with a helpful resource.

### Acronyms

**BERT** Bidirectional Encoder Representation From Transformers.

**BioBERT** Bidirectional Encoder Representations from Transformers for Biomedical Text Mining.

**BLUE** Biomedical Language Understanding Evaluation.

**C4** Colossal Clean Crawled Corpus.

**LLaMA** Large Language Model Meta AI.

**LLM** Large Language Model.

**MTEB** Massive text embedding benchmark.

**RoBERTa** Robustly Optimized BERT approach.

## 6. References

1. Singh, S.; Thakur, H.K. Survey of various AI chatbots based on technology used. In Proceedings of the 2020 8th International Conference on Reliability, Infocom Technologies and Optimization (Trends and Future Directions)(ICRITO). IEEE, 2020, pp. 1074–1079.
2. Weizenbaum, J. ELIZA—a computer program for the study of natural language communication between man and machine. *Communications of the ACM* **1966**, *9*, 36–45.

3. Vaswani, A.; Shazeer, N.; Parmar, N.; Uszkoreit, J.; Jones, L.; Gomez, A.N.; Kaiser, Ł.; Polosukhin, I. Attention is all you need. *Advances in neural information processing systems* **2017**, *30*.
4. Chang, T.A.; Bergen, B.K. Language model behavior: A comprehensive survey. *arXiv preprint arXiv:2303.11504* **2023**.
5. Uszkoreit, J. Transformer: A novel neural network architecture for language understanding. *Google AI Blog* **2017**, *31*.
6. Radford, A.; Narasimhan, K.; Salimans, T.; Sutskever, I.; et al. Improving language understanding by generative pre-training **2018**.
7. Devlin, J.; Chang, M.W.; Lee, K.; Toutanova, K. Bert: Pre-training of deep bidirectional transformers for language understanding. *arXiv preprint arXiv:1810.04805* **2018**.
8. GPT-4 is OpenAI's most advanced system, producing safer and more useful responses. https://openai.com/product/gpt-4. [Online; Accessed 06-17-2023].
9. Introducing LLaMA: A foundational, 65-billion-parameter large language model. https://ai.facebook.com/blog/large-language-model-llama-meta-ai/. [Online; Accessed 06-17-2023].
10. Chowdhery, A.; Narang, S.; Devlin, J.; Bosma, M.; Mishra, G.; Roberts, A.; Barham, P.; Chung, H.W.; Sutton, C.; Gehrmann, S.; et al. Palm: Scaling language modeling with pathways. *arXiv preprint arXiv:2204.02311* **2022**.
11. Zhu, D.; Chen, J.; Shen, X.; Li, X.; Elhoseiny, M. Minigpt-4: Enhancing vision-language understanding with advanced large language models. *arXiv preprint arXiv:2304.10592* **2023**.
12. Yang, H.; Liu, X.Y.; Wang, C.D. FinGPT: Open-Source Financial Large Language Models. *arXiv preprint arXiv:2306.06031* **2023**.
13. Zhang, S.; Roller, S.; Goyal, N.; Artetxe, M.; Chen, M.; Chen, S.; Dewan, C.; Diab, M.; Li, X.; Lin, X.V.; et al. Opt: Open pre-trained transformer language models. *arXiv preprint arXiv:2205.01068* **2022**.
14. Gu, Y.; Tinn, R.; Cheng, H.; Lucas, M.; Usuyama, N.; Liu, X.; Naumann, T.; Gao, J.; Poon, H. Domain-specific language model pretraining for biomedical natural language processing. *ACM Transactions on Computing for Healthcare (HEALTH)* **2021**, *3*, 1–23.
15. Wu, S.; Irsoy, O.; Lu, S.; Dabravolski, V.; Dredze, M.; Gehrmann, S.; Kambadur, P.; Rosenberg, D.; Mann, G. Bloomberggpt: A large language model for finance. *arXiv preprint arXiv:2303.17564* **2023**.
16. Scao, T.L.; Fan, A.; Akiki, C.; Pavlick, E.; Ilić, S.; Hesslow, D.; Castagné, R.; Luccioni, A.S.; Yvon, F.; Gallé, M.; et al. Bloom: A 176b-parameter open-access multilingual language model. *arXiv preprint arXiv:2211.05100* **2022**.
17. Black, S.; Biderman, S.; Hallahan, E.; Anthony, Q.; Gao, L.; Golding, L.; He, H.; Leahy, C.; McDonell, K.; Phang, J.; et al. Gpt-neox-20b: An open-source autoregressive language model. *arXiv preprint arXiv:2204.06745* **2022**.
18. Liu, Y.; Ott, M.; Goyal, N.; Du, J.; Joshi, M.; Chen, D.; Levy, O.; Lewis, M.; Zettlemoyer, L.; Stoyanov, V. Roberta: A robustly optimized bert pretraining approach. *arXiv preprint arXiv:1907.11692* **2019**.
19. EDWARDS, B. A really big deal"—Dolly is a free, open source, ChatGPT-style AI model. https://arstechnica.com/information-technology/2023/04/a-really-big-deal-dolly-is-a-free-open-source-chatgpt-style-ai-model/, 2023. [Online; Accessed 06-17-2023].
20. Brown, T.; Mann, B.; Ryder, N.; Subbiah, M.; Kaplan, J.D.; Dhariwal, P.; Neelakantan, A.; Shyam, P.; Sastry, G.; Askell, A.; et al. Language models are few-shot learners. *Advances in neural information processing systems* **2020**, *33*, 1877–1901.
21. Peng, B.; Li, C.; He, P.; Galley, M.; Gao, J. Instruction tuning with gpt-4. *arXiv preprint arXiv:2304.03277* **2023**.
22. Lopez-Lira, A.; Tang, Y. Can chatgpt forecast stock price movements? Return predictability and large language models. *arXiv preprint arXiv:2304.07619* **2023**.
23. Lee, A. What are large language models used for. *NVIDIA Blog* **2023**.
24. Zhao, W.X.; Zhou, K.; Li, J.; Tang, T.; Wang, X.; Hou, Y.; Min, Y.; Zhang, B.; Zhang, J.; Dong, Z.; et al. A survey of large language models. *arXiv preprint arXiv:2303.18223* **2023**.
25. Gao, J.; Lin, C.Y. Introduction to the special issue on statistical language modeling, 2004.
26. Melis, G.; Dyer, C.; Blunsom, P. On the state of the art of evaluation in neural language models. *arXiv preprint arXiv:1707.05589* **2017**.
27. Bengio, Y.; Ducharme, R.; Vincent, P. A neural probabilistic language model. *Advances in neural information processing systems* **2000**, *13*.

28. Yang, X.; Chen, A.; PourNejatian, N.; Shin, H.C.; Smith, K.E.; Parisien, C.; Compas, C.; Martin, C.; Costa, A.B.; Flores, M.G.; et al. A large language model for electronic health records. *npj Digital Medicine* **2022**, *5*, 194.
29. Lee, J.; Yoon, W.; Kim, S.; Kim, D.; Kim, S.; So, C.H.; Kang, J. BioBERT: a pre-trained biomedical language representation model for biomedical text mining. *Bioinformatics* **2020**, *36*, 1234–1240.
30. Alsentzer, E.; Murphy, J.R.; Boag, W.; Weng, W.H.; Jin, D.; Naumann, T.; McDermott, M. Publicly available clinical BERT embeddings. *arXiv preprint arXiv:1904.03323* **2019**.
31. Shin, H.C.; Zhang, Y.; Bakhturina, E.; Puri, R.; Patwary, M.; Shoeybi, M.; Mani, R. BioMegatron: Larger biomedical domain language model. *arXiv preprint arXiv:2010.06060* **2020**.
32. Yang, X.; Chen, A.; PourNejatian, N.; Shin, H.C.; Smith, K.E.; Parisien, C.; Compas, C.; Martin, C.; Flores, M.G.; Zhang, Y.; et al. Gatortron: A large clinical language model to unlock patient information from unstructured electronic health records. *arXiv preprint arXiv:2203.03540* **2022**.
33. LEE, A. What Are Large Language Models Used For? https://blogs.nvidia.com/blog/2023/01/26/what-are-large-language-models-used-for/, 2023. [Online; Accessed 06-17-2023].
34. BigScience Blog. https://bigscience.huggingface.co/blog/bloom, 2023. [Online; Accessed 06-17-2023].
35. Lee, S.; Kim, W.J.; Ye, J.C. LLM Itself Can Read and Generate CXR Images. *arXiv preprint arXiv:2305.11490* **2023**.
36. Jelinek, F. *Statistical methods for speech recognition*; MIT press, 1998.
37. Rosenfeld, R. Two decades of statistical language modeling: Where do we go from here? *Proceedings of the IEEE* **2000**, *88*, 1270–1278.
38. Peters, M.E.; Neumann, M.; Iyyer, M.; Gardner, M.; Clark, C.; Lee, K.; Zettlemoyer, L. "Deep contextualized word representations. *In Proceedings of the 2018 Conference of the North American Chapter of the Association for Computational Linguistics: Human Language Technologies, NAACL-HLT 2018, New Orleans, Louisiana,USA*.
39. Kaplan, J.; McCandlish, S.; Henighan, T.; Brown, T.B.; Chess, B.; Child, R.; Gray, S.; Radford, A.; Wu, J.; Amodei, D. Scaling laws for neural language models. *arXiv preprint arXiv:2001.08361* **2020**.
40. Bengio, Y.; Senécal, J.S. Adaptive importance sampling to accelerate training of a neural probabilistic language model. Technical report, IDIAP, 2003.
41. Chen, X.; Wang, X.; Changpinyo, S.; Piergiovanni, A.; Padlewski, P.; Salz, D.; Goodman, S.; Grycner, A.; Mustafa, B.; Beyer, L.; et al. Pali: A jointly-scaled multilingual language-image model. *arXiv preprint arXiv:2209.06794* **2022**.
42. Scheible, R.; Thomczyk, F.; Tippmann, P.; Jaravine, V.; Boeker, M. GottBERT: a pure German language model. *arXiv preprint arXiv:2012.02110* **2020**.
43. Soltan, S.; Ananthakrishnan, S.; FitzGerald, J.; Gupta, R.; Hamza, W.; Khan, H.; Peris, C.; Rawls, S.; Rosenbaum, A.; Rumshisky, A.; et al. Alexatm 20b: Few-shot learning using a large-scale multilingual seq2seq model. *arXiv preprint arXiv:2208.01448* **2022**.
44. Dossou, B.F.; Tonja, A.L.; Yousuf, O.; Osei, S.; Oppong, A.; Shode, I.; Awoyomi, O.O.; Emezue, C.C. AfroLM: A Self-Active Learning-based Multilingual Pretrained Language Model for 23 African Languages. *arXiv preprint arXiv:2211.03263* **2022**.
45. Li, Y.; Mamouei, M.; Salimi-Khorshidi, G.; Rao, S.; Hassaine, A.; Canoy, D.; Lukasiewicz, T.; Rahimi, K. Hi-BEHRT: Hierarchical Transformer-based model for accurate prediction of clinical events using multimodal longitudinal electronic health records. *IEEE Journal of Biomedical and Health Informatics* **2022**.
46. Wang, X.; Xu, X.; Tong, W.; Roberts, R.; Liu, Z. InferBERT: a transformer-based causal inference framework for enhancing pharmacovigilance. *Frontiers in Artificial Intelligence* **2021**, *4*, 659622.
47. Arora, A.; Arora, A. The promise of large language models in health care. *The Lancet* **2023**, *401*, 641.
48. Santos, T.; Tariq, A.; Das, S.; Vayalpati, K.; Smith, G.H.; Trivedi, H.; Banerjee, I. PathologyBERT–Pre-trained Vs. A New Transformer Language Model for Pathology Domain. *arXiv preprint arXiv:2205.06885* **2022**.
49. Jaiswal, A.; Tang, L.; Ghosh, M.; Rousseau, J.F.; Peng, Y.; Ding, Y. RadBERT-CL: factually-aware contrastive learning for radiology report classification. In Proceedings of the Machine Learning for Health. PMLR, 2021, pp. 196–208.
50. Peng, Y.; Yan, S.; Lu, Z. Transfer learning in biomedical natural language processing: an evaluation of BERT and ELMo on ten benchmarking datasets. *arXiv preprint arXiv:1906.05474* **2019**.

51. Beltagy, I.; Lo, K.; Cohan, A. SciBERT: A pretrained language model for scientific text. *arXiv preprint arXiv:1903.10676* **2019**.
52. Lin, C.; Miller, T.; Dligach, D.; Bethard, S.; Savova, G. EntityBERT: Entity-centric masking strategy for model pretraining for the clinical domain. Association for Computational Linguistics (ACL), 2021.
53. Yan, A.; McAuley, J.; Lu, X.; Du, J.; Chang, E.Y.; Gentili, A.; Hsu, C.N. RadBERT: Adapting transformer-based language models to radiology. *Radiology: Artificial Intelligence* **2022**, *4*, e210258.
54. Gao, L.; Biderman, S.; Black, S.; Golding, L.; Hoppe, T.; Foster, C.; Phang, J.; He, H.; Thite, A.; Nabeshima, N.; et al. The pile: An 800gb dataset of diverse text for language modeling. *arXiv preprint arXiv:2101.00027* **2020**.
55. Araci, D. Finbert: Financial sentiment analysis with pre-trained language models. *arXiv preprint arXiv:1908.10063* **2019**.
56. Xie, Q.; Han, W.; Zhang, X.; Lai, Y.; Peng, M.; Lopez-Lira, A.; Huang, J. PIXIU: A Large Language Model, Instruction Data and Evaluation Benchmark for Finance. *arXiv preprint arXiv:2306.05443* **2023**.
57. Alayrac, J.B.; Donahue, J.; Luc, P.; Miech, A.; Barr, I.; Hasson, Y.; Lenc, K.; Mensch, A.; Millican, K.; Reynolds, M.; et al. Flamingo: a visual language model for few-shot learning. *Advances in Neural Information Processing Systems* **2022**, *35*, 23716–23736.
58. Monajatipoor, M.; Rouhsedaghat, M.; Li, L.H.; Jay Kuo, C.C.; Chien, A.; Chang, K.W. Berthop: An effective vision-and-language model for chest x-ray disease diagnosis. In Proceedings of the Medical Image Computing and Computer Assisted Intervention–MICCAI 2022: 25th International Conference, Singapore, September 18–22, 2022, Proceedings, Part V. Springer, 2022, pp. 725–734.
59. Kuo, W.; Cui, Y.; Gu, X.; Piergiovanni, A.; Angelova, A. F-VLM: Open-Vocabulary Object Detection upon Frozen Vision and Language Models. *arXiv preprint arXiv:2209.15639* **2022**.
60. Hong, Y.; Wu, Q.; Qi, Y.; Rodriguez-Opazo, C.; Gould, S. Vln bert: A recurrent vision-and-language bert for navigation. In Proceedings of the Proceedings of the IEEE/CVF conference on Computer Vision and Pattern Recognition, 2021, pp. 1643–1653.
61. Thrush, T.; Jiang, R.; Bartolo, M.; Singh, A.; Williams, A.; Kiela, D.; Ross, C. Winoground: Probing vision and language models for visio-linguistic compositionality. In Proceedings of the Proceedings of the IEEE/CVF Conference on Computer Vision and Pattern Recognition, 2022, pp. 5238–5248.
62. Wang, J.; Hu, X.; Zhang, P.; Li, X.; Wang, L.; Zhang, L.; Gao, J.; Liu, Z. Minivlm: A smaller and faster vision-language model. *arXiv preprint arXiv:2012.06946* **2020**.
63. Luo, Z.; Xu, C.; Zhao, P.; Sun, Q.; Geng, X.; Hu, W.; Tao, C.; Ma, J.; Lin, Q.; Jiang, D. WizardCoder: Empowering Code Large Language Models with Evol-Instruct. *arXiv preprint arXiv:2306.08568* **2023**.
64. Nijkamp, E.; Pang, B.; Hayashi, H.; Tu, L.; Wang, H.; Zhou, Y.; Savarese, S.; Xiong, C. Codegen: An open large language model for code with multi-turn program synthesis. *arXiv preprint arXiv:2203.13474* **2022**.
65. Jain, N.; Vaidyanath, S.; Iyer, A.; Natarajan, N.; Parthasarathy, S.; Rajamani, S.; Sharma, R. Jigsaw: Large language models meet program synthesis. In Proceedings of the Proceedings of the 44th International Conference on Software Engineering, 2022, pp. 1219–1231.
66. Wang, Y.; Wang, W.; Joty, S.; Hoi, S.C. Codet5: Identifier-aware unified pre-trained encoder-decoder models for code understanding and generation. *arXiv preprint arXiv:2109.00859* **2021**.
67. Hoffmann, J.; Borgeaud, S.; Mensch, A.; Buchatskaya, E.; Cai, T.; Rutherford, E.; Casas, D.d.L.; Hendricks, L.A.; Welbl, J.; Clark, A.; et al. Training compute-optimal large language models. *arXiv preprint arXiv:2203.15556* **2022**.
68. Chen, X.; Fang, H.; Lin, T.Y.; Vedantam, R.; Gupta, S.; Dollár, P.; Zitnick, C.L. Microsoft coco captions: Data collection and evaluation server. *arXiv preprint arXiv:1504.00325* **2015**.
69. Thapliyal, A.V.; Pont-Tuset, J.; Chen, X.; Soricut, R. Crossmodal-3600: A massively multilingual multimodal evaluation dataset. *arXiv preprint arXiv:2205.12522* **2022**.
70. Xue, L.; Constant, N.; Roberts, A.; Kale, M.; Al-Rfou, R.; Siddhant, A.; Barua, A.; Raffel, C. mT5: A massively multilingual pre-trained text-to-text transformer. *arXiv preprint arXiv:2010.11934* **2020**.
71. Laurençon, H.; Saulnier, L.; Wang, T.; Akiki, C.; Villanova del Moral, A.; Le Scao, T.; Von Werra, L.; Mou, C.; González Ponferrada, E.; Nguyen, H.; et al. The bigscience roots corpus: A 1.6 tb composite multilingual dataset. *Advances in Neural Information Processing Systems* **2022**, *35*, 31809–31826.

72. Lhoest, Q.; Villanova del Moral, A.; Jernite, Y.; Thakur, A.; von Platen, P.; Patil, S.; Chaumond, J.; Drame, M.; Plu, J.; Tunstall, L.; et al. Datasets: A Community Library for Natural Language Processing. In Proceedings of the Proceedings of the 2021 Conference on Empirical Methods in Natural Language Processing: System Demonstrations; Association for Computational Linguistics: Online and Punta Cana, Dominican Republic, 2021; pp. 175–184. https://doi.org/10.18653/v1/2021.emnlp-demo.21.
73. Liang, P.; Bommasani, R.; Lee, T.; Tsipras, D.; Soylu, D.; Yasunaga, M.; Zhang, Y.; Narayanan, D.; Wu, Y.; Kumar, A.; et al. Holistic evaluation of language models. *arXiv preprint arXiv:2211.09110* **2022**.
74. Adelani, D.I.; Alabi, J.O.; Fan, A.; Kreutzer, J.; Shen, X.; Reid, M.; Ruiter, D.; Klakow, D.; Nabende, P.; Chang, E.; et al. A few thousand translations go a long way! leveraging pre-trained models for african news translation. *arXiv preprint arXiv:2205.02022* **2022**.
75. Ogueji, K.; Zhu, Y.; Lin, J. Small Data? No Problem! Exploring the Viability of Pretrained Multilingual Language Models for Low-resourced Languages. In Proceedings of the Proceedings of the 1st Workshop on Multilingual Representation Learning; Association for Computational Linguistics: Punta Cana, Dominican Republic, 2021; pp. 116–126. https://doi.org/10.18653/v1/2021.mrl-1.11.
76. Shode, I.; Adelani, D.I.; Feldman, A. yosm: A new yoruba sentiment corpus for movie reviews. *arXiv preprint arXiv:2204.09711* **2022**.
77. Herlihy, C.; Rudinger, R. MedNLI is not immune: Natural language inference artifacts in the clinical domain. *arXiv preprint arXiv:2106.01491* **2021**.
78. Wang, Y.; Fu, S.; Shen, F.; Henry, S.; Uzuner, O.; Liu, H.; et al. The 2019 n2c2/ohnlp track on clinical semantic textual similarity: overview. *JMIR medical informatics* **2020**, *8*, e23375.
79. Pampari, A.; Raghavan, P.; Liang, J.; Peng, J. emrqa: A large corpus for question answering on electronic medical records. *arXiv preprint arXiv:1809.00732* **2018**.
80. Herrett, E.; Gallagher, A.M.; Bhaskaran, K.; Forbes, H.; Mathur, R.; Van Staa, T.; Smeeth, L. Data resource profile: clinical practice research datalink (CPRD). *International journal of epidemiology* **2015**, *44*, 827–836.
81. Conrad, N.; Judge, A.; Tran, J.; Mohseni, H.; Hedgecott, D.; Crespillo, A.P.; Allison, M.; Hemingway, H.; Cleland, J.G.; McMurray, J.J.; et al. Temporal trends and patterns in heart failure incidence: a population-based study of 4 million individuals. *The Lancet* **2018**, *391*, 572–580.
82. Kuan, V.; Denaxas, S.; Gonzalez-Izquierdo, A.; Direk, K.; Bhatti, O.; Husain, S.; Sutaria, S.; Hingorani, M.; Nitsch, D.; Parisinos, C.A.; et al. A chronological map of 308 physical and mental health conditions from 4 million individuals in the English National Health Service. *The Lancet Digital Health* **2019**, *1*, e63–e77.
83. XingqiaoWang. DeepCausalPV-master. https://github.com/XingqiaoWang/DeepCausalPV-master, 2021. [Online; Accessed 06-17-2023].
84. Johnson, A.E.; Pollard, T.J.; Greenbaum, N.R.; Lungren, M.P.; Deng, C.y.; Peng, Y.; Lu, Z.; Mark, R.G.; Berkowitz, S.J.; Horng, S. MIMIC-CXR-JPG, a large publicly available database of labeled chest radiographs. *arXiv preprint arXiv:1901.07042* **2019**.
85. Irvin, J.; Rajpurkar, P.; Ko, M.; Yu, Y.; Ciurea-Ilcus, S.; Chute, C.; Marklund, H.; Haghgoo, B.; Ball, R.; Shpanskaya, K.; et al. Chexpert: A large chest radiograph dataset with uncertainty labels and expert comparison. In Proceedings of the Proceedings of the AAAI conference on artificial intelligence, 2019, Vol. 33, pp. 590–597.
86. Li, J.; Sun, Y.; Johnson, R.J.; Sciaky, D.; Wei, C.H.; Leaman, R.; Davis, A.P.; Mattingly, C.J.; Wiegers, T.C.; Lu, Z. BioCreative V CDR task corpus: a resource for chemical disease relation extraction. *Database* **2016**, *2016*.
87. Doğan, R.I.; Leaman, R.; Lu, Z. NCBI disease corpus: a resource for disease name recognition and concept normalization. *Journal of biomedical informatics* **2014**, *47*, 1–10.
88. Smith, L.; Tanabe, L.K.; Kuo, C.J.; Chung, I.; Hsu, C.N.; Lin, Y.S.; Klinger, R.; Friedrich, C.M.; Ganchev, K.; Torii, M.; et al. Overview of BioCreative II gene mention recognition. *Genome biology* **2008**, *9*, 1–19.
89. Jin, Q.; Dhingra, B.; Liu, Z.; Cohen, W.W.; Lu, X. Pubmedqa: A dataset for biomedical research question answering. *arXiv preprint arXiv:1909.06146* **2019**.
90. Nye, B.; Li, J.J.; Patel, R.; Yang, Y.; Marshall, I.J.; Nenkova, A.; Wallace, B.C. A corpus with multi-level annotations of patients, interventions and outcomes to support language processing for medical literature. In Proceedings of the Proceedings of the conference. Association for Computational Linguistics. Meeting. NIH Public Access, 2018, Vol. 2018, p. 197.
91. Luan, Y.; He, L.; Ostendorf, M.; Hajishirzi, H. Multi-task identification of entities, relations, and coreference for scientific knowledge graph construction. *arXiv preprint arXiv:1808.09602* **2018**.
92. Jurgens, D.; Kumar, S.; Hoover, R.; McFarland, D.; Jurafsky, D. Measuring the evolution of a scientific field through citation frames. *Transactions of the Association for Computational Linguistics* **2018**, *6*, 391–406.
93. Ammar, W.; Groeneveld, D.; Bhagavatula, C.; Beltagy, I.; Crawford, M.; Downey, D.; Dunkelberger, J.; Elgohary, A.; Feldman, S.; Ha, V.; et al. Construction of the literature graph in semantic scholar. *arXiv preprint arXiv:1805.02262* **2018**.
94. Uzuner, Ö.; Luo, Y.; Szolovits, P. Evaluating the state-of-the-art in automatic de-identification. *Journal of the American Medical Informatics Association* **2007**, *14*, 550–563.
95. Uzuner, Ö.; South, B.R.; Shen, S.; DuVall, S.L. 2010 i2b2/VA challenge on concepts, assertions, and relations in clinical text. *Journal of the American Medical Informatics Association* **2011**, *18*, 552–556.
96. Sun, W.; Rumshisky, A.; Uzuner, O. Evaluating temporal relations in clinical text: 2012 i2b2 challenge. *Journal of the American Medical Informatics Association* **2013**, *20*, 806–813.
97. Stubbs, A.; Uzuner, Ö. Annotating longitudinal clinical narratives for de-identification: The 2014 i2b2/UTHealth corpus. *Journal of biomedical informatics* **2015**, *58*, S20–S29.
98. Romanov, A.; Shivade, C. Lessons from natural language inference in the clinical domain. *arXiv preprint arXiv:1808.06752* **2018**.
99. Johnson, A.E.; Pollard, T.J.; Shen, L.; Lehman, L.w.H.; Feng, M.; Ghassemi, M.; Moody, B.; Szolovits, P.; Anthony Celi, L.; Mark, R.G. MIMIC-III, a freely accessible critical care database. *Scientific data* **2016**, *3*, 1–9.
100. Krallinger, M.; Rabal, O.; Leitner, F.; Vazquez, M.; Salgado, D.; Lu, Z.; Leaman, R.; Lu, Y.; Ji, D.; Lowe, D.M.; et al. The CHEMDNER corpus of chemicals and drugs and its annotation principles. *Journal of cheminformatics* **2015**, *7*, 1–17.
101. Tsatsaronis, G.; Balikas, G.; Malakasiotis, P.; Partalas, I.; Zschunke, M.; Alvers, M.R.; Weissenborn, D.; Krithara, A.; Petridis, S.; Polychronopoulos, D.; et al. An overview of the BIOASQ large-scale biomedical semantic indexing and question answering competition. *BMC bioinformatics* **2015**, *16*, 1–28.
102. Peters, M.E.; Ammar, W.; Bhagavatula, C.; Power, R. Semi-supervised sequence tagging with bidirectional language models. *arXiv preprint arXiv:1705.00108* **2017**.
103. Raffel, C.; Shazeer, N.; Roberts, A.; Lee, K.; Narang, S.; Matena, M.; Zhou, Y.; Li, W.; Liu, P.J. Exploring the Limits of Transfer Learning with a Unified Text-to-Text Transformer. *Journal of Machine Learning Research* **2020**, *21*, 1–67.
104. Shah, R.; Chawla, K.; Eidnani, D.; Shah, A.; Du, W.; Chava, S.; Raman, N.; Smiley, C.; Chen, J.; Yang, D. When FLUE Meets FLANG: Benchmarks and Large Pretrained Language Model for Financial Domain. In Proceedings of the Proceedings of the 2022 Conference on Empirical Methods in Natural Language Processing; Association for Computational Linguistics: Abu Dhabi, United Arab Emirates, 2022; pp. 2322–2335.
105. Zhu, Y.; Kiros, R.; Zemel, R.; Salakhutdinov, R.; Urtasun, R.; Torralba, A.; Fidler, S. Aligning books and movies: Towards story-like visual explanations by watching movies and reading books. In Proceedings of the Proceedings of the IEEE international conference on computer vision, 2015, pp. 19–27.
106. Rajaraman, A.; Ullman, J.D. *Mining of massive datasets*; Cambridge University Press, 2011.
107. Wang, A.; Pruksachatkun, Y.; Nangia, N.; Singh, A.; Michael, J.; Hill, F.; Levy, O.; Bowman, S. Superglue: A stickier benchmark for general-purpose language understanding systems. *Advances in neural information processing systems* **2019**, *32*.
108. Muennighoff, N.; Tazi, N.; Magne, L.; Reimers, N. MTEB: Massive text embedding benchmark. *arXiv preprint arXiv:2210.07316* **2022**.
109. Muennighoff, N.; Wang, T.; Sutawika, L.; Roberts, A.; Biderman, S.; Scao, T.L.; Bari, M.S.; Shen, S.; Yong, Z.X.; Schoelkopf, H.; et al. Crosslingual generalization through multitask finetuning. *arXiv preprint arXiv:2211.01786* **2022**.
110. Reuters Corpora (RCV1, RCV2, TRC2). https://trec.nist.gov/data/reuters/reuters.html, 2004. [Online; Accessed 06-17-2023].
111. Malo, P.; Sinha, A.; Takala, P.; Korhonen, P.; Wallenius, J. FinancialPhraseBank-v1. 0, 2013.

112. Maia, M.; Handschuh, S.; Freitas, A.; Davis, B.; McDermott, R.; Zarrouk, M.; Balahur, A. Www'18 open challenge: financial opinion mining and question answering. In Proceedings of the Companion proceedings of the the web conference 2018, 2018, pp. 1941–1942.
113. Jia, C.; Yang, Y.; Xia, Y.; Chen, Y.T.; Parekh, Z.; Pham, H.; Le, Q.; Sung, Y.H.; Li, Z.; Duerig, T. Scaling up visual and vision-language representation learning with noisy text supervision. In Proceedings of the International conference on machine learning. PMLR, 2021, pp. 4904–4916.
114. Peng, Y.; Wang, X.; Lu, L.; Bagheri, M.; Summers, R.; Lu, Z. NegBio: a high-performance tool for negation and uncertainty detection in radiology reports. *AMIA Summits on Translational Science Proceedings* **2018**, *2018*, 188.
115. Gupta, A.; Dollar, P.; Girshick, R. Lvis: A dataset for large vocabulary instance segmentation. In Proceedings of the Proceedings of the IEEE/CVF conference on computer vision and pattern recognition, 2019, pp. 5356–5364.
116. Changpinyo, S.; Sharma, P.; Ding, N.; Soricut, R. Conceptual 12m: Pushing web-scale image-text pre-training to recognize long-tail visual concepts. In Proceedings of the Proceedings of the IEEE/CVF Conference on Computer Vision and Pattern Recognition, 2021, pp. 3558–3568.
117. Ordonez, V.; Kulkarni, G.; Berg, T. Im2text: Describing images using 1 million captioned photographs. *Advances in neural information processing systems* **2011**, *24*.
118. Schuhmann, C.; Vencu, R.; Beaumont, R.; Kaczmarczyk, R.; Mullis, C.; Katta, A.; Coombes, T.; Jitsev, J.; Komatsuzaki, A. Laion-400m: Open dataset of clip-filtered 400 million image-text pairs. *arXiv preprint arXiv:2111.02114* **2021**.
119. Zhou, L.; Palangi, H.; Zhang, L.; Hu, H.; Corso, J.; Gao, J. Unified vision-language pre-training for image captioning and vqa. In Proceedings of the Proceedings of the AAAI conference on artificial intelligence, 2020, Vol. 34, pp. 13041–13049.
120. Goyal, Y.; Khot, T.; Summers-Stay, D.; Batra, D.; Parikh, D. Making the v in vqa matter: Elevating the role of image understanding in visual question answering. In Proceedings of the Proceedings of the IEEE conference on computer vision and pattern recognition, 2017, pp. 6904–6913.
121. Suhr, A.; Zhou, S.; Zhang, A.; Zhang, I.; Bai, H.; Artzi, Y. A corpus for reasoning about natural language grounded in photographs. *arXiv preprint arXiv:1811.00491* **2018**.
122. Anderson, P.; Wu, Q.; Teney, D.; Bruce, J.; Johnson, M.; Sünderhauf, N.; Reid, I.; Gould, S.; Van Den Hengel, A. Vision-and-language navigation: Interpreting visually-grounded navigation instructions in real environments. In Proceedings of the Proceedings of the IEEE conference on computer vision and pattern recognition, 2018, pp. 3674–3683.
123. Qi, Y.; Wu, Q.; Anderson, P.; Wang, X.; Wang, W.Y.; Shen, C.; Hengel, A.v.d. Reverie: Remote embodied visual referring expression in real indoor environments. In Proceedings of the Proceedings of the IEEE/CVF Conference on Computer Vision and Pattern Recognition, 2020, pp. 9982–9991.
124. Chang, A.; Dai, A.; Funkhouser, T.; Halber, M.; Niessner, M.; Savva, M.; Song, S.; Zeng, A.; Zhang, Y. Matterport3D: Learning from RGB-D Data in Indoor Environments. *International Conference on 3D Vision (3DV)* **2017**.
125. Chen, M.; Tworek, J.; Jun, H.; Yuan, Q.; Pinto, H.P.d.O.; Kaplan, J.; Edwards, H.; Burda, Y.; Joseph, N.; Brockman, G.; et al. Evaluating large language models trained on code. *arXiv preprint arXiv:2107.03374* **2021**.
126. Liu, J.; Xia, C.S.; Wang, Y.; Zhang, L. Is your code generated by chatgpt really correct? rigorous evaluation of large language models for code generation. *arXiv preprint arXiv:2305.01210* **2023**.
127. Austin, J.; Odena, A.; Nye, M.; Bosma, M.; Michalewski, H.; Dohan, D.; Jiang, E.; Cai, C.; Terry, M.; Le, Q.; et al. Program synthesis with large language models. *arXiv preprint arXiv:2108.07732* **2021**.
128. Lai, Y.; Li, C.; Wang, Y.; Zhang, T.; Zhong, R.; Zettlemoyer, L.; Yih, S.W.t.; Fried, D.; Wang, S.; Yu, T. DS-1000: A natural and reliable benchmark for data science code generation. *arXiv preprint arXiv:2211.11501* **2022**.
129. Xu, C.; Sun, Q.; Zheng, K.; Geng, X.; Zhao, P.; Feng, J.; Tao, C.; Jiang, D. Wizardlm: Empowering large language models to follow complex instructions. *arXiv preprint arXiv:2304.12244* **2023**.
130. Husain, H.; Wu, H.H.; Gazit, T.; Allamanis, M.; Brockschmidt, M. Codesearchnet challenge: Evaluating the state of semantic code search. *arXiv preprint arXiv:1909.09436* **2019**.
131. Radford, A.; Wu, J.; Child, R.; Luan, D.; Amodei, D.; Sutskever, I.; et al. Language models are unsupervised multitask learners. *OpenAI blog* **2019**, *1*, 9.
132. Koubaa, A. GPT-4 vs. GPT-3.5: A concise showdown **2023**.
133. GPT-3. https://en.wikipedia.org/wiki/GPT-3#GPT-3.5, 2023. [Online; Accessed 06-17-2023].

134. Zhang, Y.; Sun, S.; Galley, M.; Chen, Y.C.; Brockett, C.; Gao, X.; Gao, J.; Liu, J.; Dolan, B. Dialogpt: Large-scale generative pre-training for conversational response generation. *arXiv preprint arXiv:1911.00536* **2019**.
135. Microsoft Research-DialoGPT. https://www.microsoft.com/en-us/research/project/large-scale-pretraining-for-response-generation/, 2019. [Online; Accessed 06-17-2023].
136. Padmanabha, N.H. A step-by-step guide to running Vicuna-13B Large Language Model on your GPU / CPU machine. https://www.linkedin.com/pulse/step-by-step-guide-running-vicuna-13b-large-language-nischal/, 2023. [Online; Accessed 06-17-2023].
137. Lieber, O.; Sharir, O.; Lenz, B.; Shoham, Y. Jurassic-1: Technical details and evaluation. *White Paper. AI21 Labs* **2021**, *1*.
138. Zeng, A.; Liu, X.; Du, Z.; Wang, Z.; Lai, H.; Ding, M.; Yang, Z.; Xu, Y.; Zheng, W.; Xia, X.; et al. Glm-130b: An open bilingual pre-trained model. *arXiv preprint arXiv:2210.02414* **2022**.
139. WIKIPEDIA. https://en.wikipedia.org/wiki/PaLM, 2022. [Online; Accessed 06-17-2023].
140. Khadhraoui, M.; Bellaaj, H.; Ammar, M.B.; Hamam, H.; Jmaiel, M. Survey of BERT-base models for scientific text classification: COVID-19 case study. *Applied Sciences* **2022**, *12*, 2891.
141. Smith, S.; Patwary, M.; Norick, B.; LeGresley, P.; Rajbhandari, S.; Casper, J.; Liu, Z.; Prabhumoye, S.; Zerveas, G.; Korthikanti, V.; et al. Using deepspeed and megatron to train megatron-turing nlg 530b, a large-scale generative language model. *arXiv preprint arXiv:2201.11990* **2022**.
142. ClinicalBERT. https://huggingface.co/medicalai/ClinicalBERT. [Online; Accessed 06-17-2023].
143. Hugging Face. https://huggingface.co/UFNLP/gatortron-base. [Online; Accessed 06-17-2023].
144. Alford, A. Google Trains 280 Billion Parameter AI Language Model Gopher. https://www.infoq.com/news/2022/01/deepmind-gopher/, 2022. [Online; Accessed 06-17-2023].
145. Hoffmann, J.; Borgeaud, S.; Mensch, A.; Buchatskaya, E.; Cai, T.; Rutherford, E.; de Las Casas, D.; Hendricks, L.A.; Welbl, J.; Clark, A.; et al. An empirical analysis of compute-optimal large language model training. *Advances in Neural Information Processing Systems* **2022**, *35*, 30016–30030.